\documentclass[runningheads]{llncs}
\usepackage[T1]{fontenc}
\usepackage{graphicx}
\usepackage{amssymb,siunitx}
\usepackage{xcolor}
\usepackage{tikz, pgfplots, algorithmic, algorithm, url, amsmath}
\usepackage{array}
\usepackage{hyperref}

\begin{document}
\title{Binary Noise for Binary Tasks: Masked Bernoulli Diffusion for Unsupervised Anomaly Detection}
\titlerunning{Binary Noise for Binary Tasks}

\author{Julia Wolleb \and
Florentin Bieder \and
Paul Friedrich  \and
Peter Zhang \and \\
Alicia Durrer \and
Philippe C. Cattin }
\authorrunning{J. Wolleb et al.}
\institute{Department of Biomedical Engineering, University of Basel, Allschwil, Switzerland\\
\email{julia.wolleb@unibas.ch}}

\maketitle              % typeset the header of the contribution
\begin{abstract}
The high performance of denoising diffusion models for image generation has paved the way for their application in unsupervised medical anomaly detection.
As diffusion-based methods require a lot of GPU memory and have long sampling times, we present a novel and fast unsupervised anomaly detection approach based on latent Bernoulli diffusion models. We first apply an autoencoder to compress the input images into a binary latent representation. Next, a diffusion model that follows a Bernoulli noise schedule is employed to this latent space and trained to restore binary latent representations from perturbed ones. The binary nature of this diffusion model allows us to identify entries in the latent space that have a high probability of flipping their binary code during the denoising process, which indicates out-of-distribution data. We propose a masking algorithm based on these probabilities, which improves the anomaly detection scores. We achieve state-of-the-art performance compared to other diffusion-based unsupervised anomaly detection algorithms while significantly reducing sampling time and memory consumption. The code is available at \url{https://github.com/JuliaWolleb/Anomaly_berdiff}.

\keywords{Bernoulli Diffusion  \and Unsupervised Anomaly Detection }
\end{abstract}
\section{Introduction}
The recent success of denoising diffusion models \cite{dhariwal2021diffusion,ho2020denoising} has paved the way for unsupervised anomaly detection methods, which aim to identify unusual patterns, outliers, or deviations from normal behavior without relying on labeled examples \cite{gonzalez2023sano,huijben2023histogram,naval2024disyre}. In this work, we aim to identify pixel-level anomalous changes in medical images while training on a dataset of healthy subjects only. 
We thus aim to transform pathological tissue of the input image into healthy tissue, and use the difference map between input and output to highlight abnormal areas {\cite{behrendt2024patched,kascenas2022denoising,pinaya2022fast,wyatt2022anoddpm}.
As approaches operating in the image-space suffer from long sampling times, we propose a novel method that relies on binary latent diffusion models \cite{wang2023binary}.  An overview of the model training is presented in Fig. \ref{fig:overview}. 
Our approach uses a binarizing autoencoder to compress input images into a binary latent space. A diffusion model that follows a Bernoulli noise schedule is then trained to gradually denoise perturbed representations within the latent space.
For each entry of the latent representation, the diffusion model directly extracts the probability of flipping its binary code during the denoising process. We argue that entries associated with anomalous changes are more likely to flip, indicating out-of-distribution data.
Based on this assumption, we design a novel masking strategy that enables reconstructions close to the input image, thus improving the accuracy of the difference map and enhancing anomaly detection scores.
We assess our method on the BRATS2020 dataset for brain tumor detection, and the OCT2017 dataset for drusen detection in retinal OCT images.

\begin{figure}[t]
\centering
\includegraphics[width=0.97\textwidth]{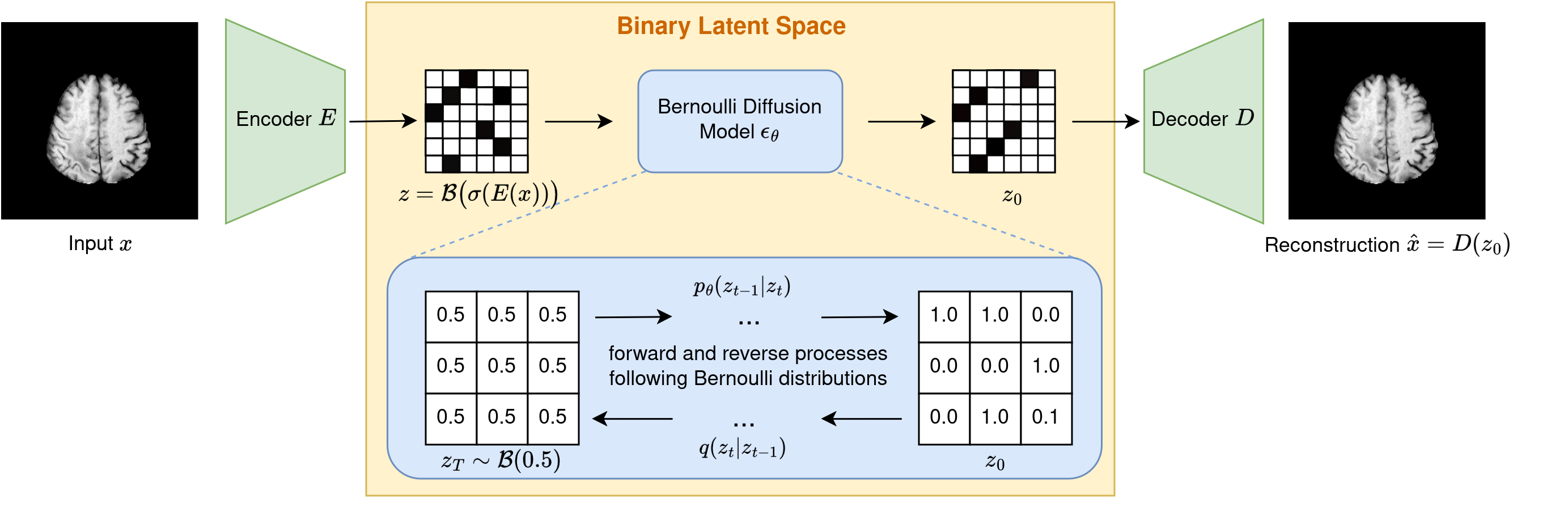}
\caption{We train a binarizing autoencoder consisting of an encoder $E$ and a decoder $D$. In the latent space, a diffusion model is trained to restore the binary latent code of healthy images by reversing a Bernoulli noise driven diffusion process. Both the autoencoder and the diffusion model are exclusively trained on healthy data.} \label{fig:overview}
\end{figure}

\subsubsection{Related Work}

Many unsupervised anomaly detection algorithms rely on variational autoencoders (VAEs) \cite{kingma} trained on healthy samples only. They extract anomaly scores from the reconstruction error between the pathological input image and its healthy reconstruction \cite{marimont2021anomaly,zimmerer2019unsupervised}. Others are based on self-supervised learning using segmentation \cite{marimont2023achieving}, patch-based regression \cite{bieder2022position}, or denoising autoencoders \cite{kascenas2022denoising}. 
Due to their high reconstruction quality, diffusion models \cite{dhariwal2021diffusion,ho2020denoising} have been applied to weakly supervised anomaly detection tasks \cite{sanchez2022healthy,wolleb2022diffusion}. If no labels are available, \textit{AnoDDPM} \cite{wyatt2022anoddpm} suggests replacing the Gaussian noise scheme with simplex noise. This is further explored in \cite{kascenas2023role}, where a coarse noise schedule is proposed. Various diffusion-based masking mechanisms have been explored \cite{behrendt2024patched,iqbal2023unsupervised,liang2023modality} to reconstruct healthy tissue.
To further improve the quality of these healthy reconstructions, \textit{AutoDDPM} \cite{bercea2023mask} proposes an iterative masking, stitching, and resampling algorithm. 
To address concerns of memory consumption and sampling time, latent diffusion models (LDMs) \cite{pinaya2022fast} encode input images into a low-dimensional latent space, where a diffusion model is trained to restore healthy latent representations.
While classical denoising diffusion models are based on Gaussian noise, the use of Bernoulli noise was introduced as early as 2015 \cite{sohl2015deep} and has been applied for fully supervised lesion segmentation \cite{chen2023berdiff}. 
Furthermore, a binary LDM, using a binarizing autoencoder in combination with a latent Bernoulli diffusion model, was proposed for image synthesis  \cite{wang2023binary}.

\subsubsection{Contribution} 
We present a novel approach for unsupervised anomaly detection in medical images. Our method uses a binary latent diffusion model based on Bernoulli noise to learn a robust representation from healthy images. 
Taking advantage of the binary nature of Bernoulli diffusion models, we design a novel masking scheme to preserve the anatomical information of the input images, thereby improving the anomaly detection performance. We achieve state-of-the-art anomaly detection results on two different medical datasets while reducing sampling time and memory footprint.

\section{Method}
Our approach consists of two neural networks, shown in Fig. \ref{fig:overview}: An autoencoder (in green) mapping an input image into a binary latent space, as well as a Bernoulli diffusion model (in blue) in the binary latent space. First, we train the autoencoder, consisting of an encoder $E$ and a decoder $D$.

\subsubsection{Autoencoder}
Following \cite{wang2023binary}, the convolutional encoder $E$ extracts a compressed representation $y=\sigma(E(x)) \in \mathbb{R}^{C \times \frac{h}{k} \times\frac{w}{k}} $ from an input image ${x \in \mathbb{R}^{c \times h \times w}.}$ The channel dimension of the input image is denoted by $c$, $k$ is the compression rate, $\sigma$ is the sigmoid activation function, and $C$ is the number of encoded feature channels.
By sampling $z \sim \mathcal{B}(y)$, we obtain a binary representation $z$. The decoder $D$ reconstructs the image $\hat{x}=D(z)$. 
We train the autoencoder using a combined MSE, perceptual, and adversarial loss between $x$ and $\hat{x}$ \cite{wang2023binary}.

\subsubsection{Bernoulli Diffusion Model}
Following \cite{chen2023berdiff}, the Bernoulli diffusion model takes the latent representation $z=z_0$ as input. The Bernoulli forward process follows a sequence of noise perturbations

\begin{equation}\label{eq:forward}
q(z_{t} \mid z_{t-1}):=\mathcal{B}\Bigl((1-\beta _{t})z_{t-1}+\frac{\beta _{t}}{2}\Bigr) \textrm{ for } t \in {1, \ldots ,T},
\end{equation}
where $\mathcal{B}$ denotes the Bernoulli distribution, and $\beta_{1:T}$ follow a predefined noise schedule. By defining $\alpha _{t}:=1-\beta _{t}$ and $\bar{\alpha}_{t}:=\prod_{s=1}^t \alpha _{s}$, we can directly compute

\begin{equation}\label{eq:property}
z_{t} = z_0 \oplus \epsilon, \qquad \textrm{ with }  \epsilon \sim \mathcal{B}\biggl(\frac{1-\bar{\alpha}_t}{2}\biggr),
\end{equation}
where $\oplus$ is the logical XOR operation.
The Bernoulli posterior is defined as
\begin{equation}
\label{eq:denoising}
q(z_{t-1} \vert z_t, z_0)=\mathcal{B}\bigl(\theta_{post}(z_t, z_0)\bigr),
\end{equation}
which is described in detail in the Supplementary Material.
For the reverse diffusion process, we train a deep neural network $\epsilon_{\theta}$, following the architecture proposed in  \cite{dhariwal2021diffusion}, to predict the residual $\hat{\epsilon}=\epsilon_{\theta}(z_t,t)$, i.e., the flipping probability of the binary code $z_t$ \cite{wang2023binary}. 
For each time step $t$ we predict $\tilde{z}_0=\lvert z_t - \epsilon_{\theta}(z_t,t)\rvert$ and approximate  $z_0$ in Equation \ref{eq:denoising}  with  $\tilde{z}_0$, i.e.,
\begin{equation}\label{eq:reverse}
p_{\theta}(z_{t-1}\vert z_t, \tilde{z_0}):= \mathcal{B}\bigl(\theta_{post}(z_t,\tilde{z}_0)\bigr) = \mathcal{B}\bigl(\theta_{post}(z_t,\lvert z_t - \epsilon_{\theta}(z_t,t)\rvert)\bigr) .
\end{equation}
To train the model $\epsilon_{\theta}$, we use the binary cross entropy (BCE) as loss function
\begin{equation}
\label{eq:loss}
\mathcal{L}(z_t,t) =BCE \bigl(\epsilon_{\theta} (z_t,t), z_t \oplus z_0\bigr).
\end{equation}
To generate synthetic images, we start from random noise $z_T \sim \mathcal{B}(\frac{1}{2})$ and iterate through the denoising process by sampling 
\begin{equation}
\label{eq:sampling}
%z_{t-1} \sim \mathcal{B}(\theta_{post}(z_t,\lvert z_t - \epsilon_{\theta}(z_t,t) \rvert) \textrm{ for } t = T, \ldots ,1.
z_{t-1} \sim \mathcal{B}(\theta_{post}(z_t,  \tilde{z}_0))\textrm{ for } t = T, \ldots ,1.
\end{equation} 

 \subsubsection{Anomaly Detection during Inference}
An overview of the pipeline is shown in Fig. \ref{fig:masking}.
To detect anomalies in an image $x$ of a possibly diseased subject, we first pass it through the encoder $E$ to get a binary representation ${z=\mathcal{B}(\sigma(E(x))}$.
Similar to other diffusion-based reconstruction methods \cite{kascenas2023role,pinaya2022fast,wolleb2022diffusion,wyatt2022anoddpm}, we add $L$ steps of noise to $z$ using Equation \ref{eq:property}. The resulting noisy latent representation $z_L$ is then passed through the denoising process described in Equation \ref{eq:sampling} for $t=L, \ldots ,1$. Since the diffusion model was trained on healthy subjects only, it restores a latent representation $z_0$ following the learned healthy distribution. We finally reconstruct an output image $\hat{x}=D(z_0)$. The desired pixel-wise anomaly map is computed by $a=\sum\limits_{c} (x-\hat{x})^2$. 

\begin{figure}[t]
\centering
\includegraphics[width=\textwidth]{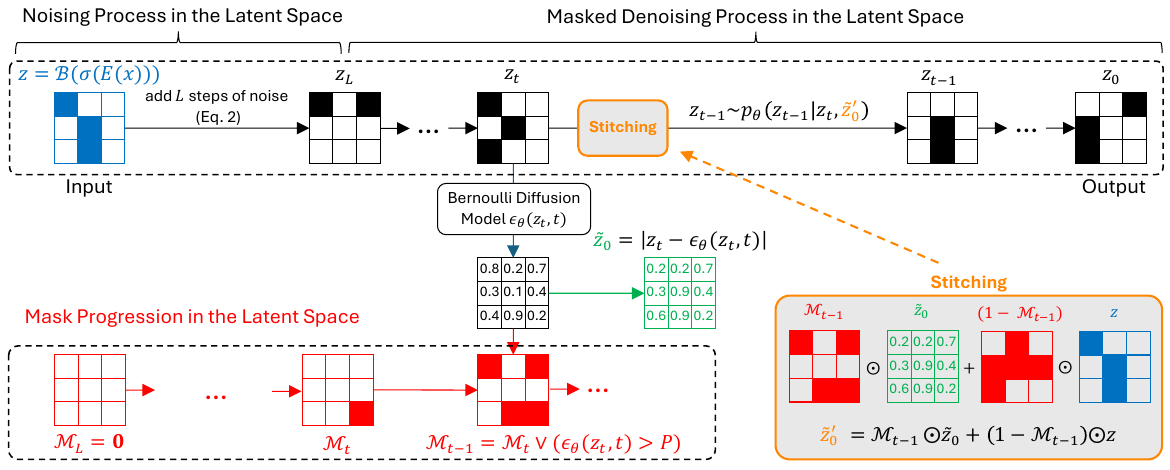}
\caption{In our proposed denoising process, we aim to restore a healthy representation $z_0$ from an initial input representation $z$. We define a mask $\mathcal{M}$ based on the threshold probability $P$. By stitching the sampled and original latent representations at each timestep, we ensure the preservation of anatomical information from the input.} \label{fig:masking}
\end{figure}

\subsubsection{Masking}

As described in \cite{bercea2023mask}, restoration-based methods face the \textit{noise paradox}: We have to strike a balance between sufficiently large noise levels $L$ while preserving identity information in healthy tissue. While increasing the noise level $L$ reduces anomalies in the reconstruction, the integrity of healthy tissue is at risk, leading to false positive pixels in the anomaly map $a$.
We propose a novel masking algorithm, presented in Fig. \ref{fig:masking}, to tackle this challenge.
We take advantage of the binary nature of our Bernoulli diffusion model. We argue that in any timestep $t$, entries of the latent representation related to anomalous changes are likely to have a high flipping probability $\epsilon_{\theta}(z_t,t)$, as they do not follow the expected healthy distribution.
First, we initialize a mask $\mathcal{M}_L=\mathbf{0}=[0]^{C \times \frac{h}{k} \times\frac{w}{k}}$ in the latent space, and define a threshold probability $P$. In every step $t$, we update $\mathcal{M}_t$ by masking out entries that have a flipping probability higher than $P$, by computing 
 $\mathcal{M}_{t-1} := \mathcal{M}_{t}  \lor (\epsilon_{\theta}(z_t,t)>P)$.
To preserve anatomical information from the original latent representation $z$, we stitch the predicted $\tilde {z}_0=\lvert z_t - \epsilon_{\theta}(z_t,t) \rvert$ and the original $z$ according to 
\begin{equation}
\tilde{z}'_0 = \mathcal{M}_{t-1} \odot \tilde{z}_0  + (1-\mathcal{M}_{t-1}) \odot z, 
\label{eq:stitching}
\end{equation}
where $\odot$ denotes an element-wise multiplication.
Thereby, we only change entries with a high flipping probability, 
while preserving the rest of the original representation $z$. 
Finally, we sample $z_{t-1} \sim \mathcal{B}(\theta_{post}(z_t, \tilde{z}'_0))$ according to Equation \ref{eq:sampling}.
The whole inference procedure is presented in Algorithm \ref{alg1}.
\begin{algorithm}
    \caption{Anomaly detection using our masked Bernoulli diffusion model}
    \label{alg1}
    \begin{algorithmic}
        \STATE \textbf{Input}: input image $x$, probability threshold $P$, noise level $L$\\
        \textbf{Output}: healthy reconstruction $\hat{x}$, anomaly map $a$\\
        \textbf{Initialize}  $\mathcal{M}_L=\mathbf{0}$, $z=\mathcal{B}(\sigma(E(x)))$\ \\
            $z_{L} \gets z \oplus \epsilon_L, \textrm{ where }  \epsilon_L \sim \mathcal{B}(\frac{1-\bar{\alpha}_L}{2})$\\
          
            \FORALL {$t$ from $L$ to $1$ }
            \STATE $\tilde{z}_0 \gets \lvert z_t - \epsilon_{\theta}(z_t,t) \rvert$
             \COMMENT {prediction of z at time 0}
          %  \STATE $ \mathcal{M}[\epsilon_{\theta}(z_t,t)>P]=1$
            \STATE $\mathcal{M}_{t-1} \gets  \mathcal{M}_t \lor (\epsilon_{\theta}(z_t,t)>P)$
            \COMMENT {update mask}
            \STATE $\tilde{z}'_0 \gets  \mathcal{M}_{t-1} \odot \tilde{z}_0 + (1- \mathcal{M}_{t-1}) \odot z $
             \COMMENT { element-wise stitching of $z$ and  $\bar{z}_0$}
            \STATE $z_{t-1} \sim \mathcal{B}(\theta_{post}(z_t,\tilde{z}'_0)) $
             \COMMENT { sample $z_{t-1} $}
            \ENDFOR
        \STATE $\hat{x}=D(z_0)$
        \STATE $a \gets \sum\limits_{c} (x-\hat{x})^2$
        \RETURN $\hat{x}$, $a$
    \end{algorithmic}
\end{algorithm}

\section{Experiments}
The binary autoencoder is trained as proposed in \cite{wang2023binary}, with a batch size of $6$. We encode images of resolution $ c \times 256 \times 256$ to a resolution of $128 \times 32 \times 32$ in the binary latent space, where $c$ is the number of input channels. For the Bernoulli diffusion model, we choose a U-Net architecture as described in \cite{chen2023berdiff,dhariwal2021diffusion}. 
The diffusion model has $T=1000$ timesteps and is trained using the Adam optimizer with a learning rate of ${10^{-4}}$ and a batch size of $32$.
By choosing the number of channels in the first layer of the diffusion model to be $128$, and using one attention head at resolution $16$, the total number of parameters is 36,034,432} for the Bernoulli diffusion model and 4,256,808 for the binarizing autoencoder. The diffusion model is trained for \num{2e5} iterations and the autoencoder for \num{1.2e4} iterations on an NVIDIA Quadro RTX 6000 GPU using Pytorch 2.1.0.
%, with Python version 3.11.5. 

\subsubsection{BRATS2020}
This dataset \cite{brats2,brats3,brats1} consists of 3D brain MR images of subjects diagnosed with a brain tumor, together with pixel-wise ground truth labels outlining tumor regions. Our focus on a 2D approach leads us to analyse axial slices. Each slice has four channels corresponding to four different MR sequences, is zero-padded to dimensions of $256 \times 256$ and normalized in the range 0 to 1. Given the typical occurrence of tumors in the central regions of the brain \cite{larjavaara2007incidence}, we omit the bottom 80 and the top 26 slices. A slice is classified as healthy if the ground truth label mask is zero. We split the dataset at the patient level, resulting in a training set of 5,598 healthy slices. The test set contains 1,038 slices with tumors and 705 without. To post-process the obtained anomaly maps $a$, similar to \cite{behrendt2024patched}, we apply median filtering with a kernel-size of 5, apply a threshold at $0.5$ to obtain a segmentation \cite{wyatt2022anoddpm}, and remove components below 10 pixels.

\subsubsection{OCT2017}
This dataset \cite{kermany2018labeled} contains optical coherence tomography (OCT) images of the retina. We select 18,232 healthy subjects for training. The dataset comprises grayscale images resized to a resolution of $256 \times 256$ and normalized to values between $0$ and $1$. 
We randomly select subjects suffering from drusen for qualitative comparison. Drusen manifest as accumulations of extracellular material between the Bruch's membrane and the retinal pigment epithelium (RPE), causing elevations of the RPE \cite{DESPOTOVIC201816}.

\section{Results and Discussion}

\subsubsection{Hyperparameter Selection}
To explore the effect of the threshold probability~$P$ and the noise level $L$, we perform a grid search on the BRATS2020 dataset. In Fig. \ref{fig:hyper} (left), we report the Dice score between the anomaly map $a$ and the ground truth tumor segmentation. Note that a threshold of $P=0$ corresponds to no mask being applied.
Notably, a noise level of $L=200$ combined with $P=0.5$ achieves the highest scores. We choose this setting for further comparison with the state-of-the-art. Additional analyses of the effect of the hyperparameters $L$ and $P$ on the restored images $\hat{x}$ can be found in the Supplementary Material.

\subsubsection{Image-Level Masking Scores}
\begin{figure}[h]
\centering
\begin{minipage}[t]{0.45\textwidth}
\vspace{0pt} 
{\resizebox{1\textwidth}{!}{% This file was created by tikzplotlib v0.9.6.
\begin{tikzpicture}

\definecolor{color0}{rgb}{0.133333333333333,0.545098039215686,0.133333333333333}
\definecolor{color1}{rgb}{1,0.647058823529412,0}

\begin{axis}[
legend cell align={left},
legend style={fill opacity=0.8, draw opacity=1, text opacity=1, at={(0.91,0.2)}, anchor=east, draw=white!80!black},
tick align=outside,
tick pos=left,
x grid style={white!69.0196078431373!black},
xlabel={Threshold value $P$ for the masking schedule},
xmin=0.275, xmax=0.825,
xtick style={color=black},
xtick={0.2,0.3,0.4,0.5,0.6,0.7,0.8,0.9},
xticklabels={0.2,no mask ,0.4,0.5,0.6,0.7,0.8,0.9},
y grid style={white!69.0196078431373!black},
ylabel={average Dice score},
ymin=0.386725, ymax=0.599575,
ytick style={color=black}
]
\addplot [semithick, color1, mark=triangle*, mark size=3, mark options={solid,rotate=180}]
table {%
0.3 0.5356
0.4 0.5137245
0.5 0.5145
0.6 0.5155
0.7 0.5173
0.8 0.51993
};
\addlegendentry{L=100}
\addplot [semithick, blue!50!black, mark=*, mark size=3, mark options={solid}]
table {%
0.3 0.50263
0.4 0.519257
0.5 0.5899
0.6 0.5843
0.7 0.56977
0.8 0.51289
};
\addlegendentry{L=200}
\addplot [semithick, color0, mark=x, mark size=3, mark options={solid}]
table {%
0.3 0.3964
0.4 0.4805
0.5 0.53245
0.6 0.541263
0.7 0.54702
0.8 0.553115
};
\addlegendentry{L=400}
\addplot [semithick, red!70!black, mark=diamond*, mark size=3, mark options={solid}]
table {%
0.3 0.440652
0.4 0.54095
0.5 0.55286
0.6 0.56019
0.7 0.56458
0.8 0.561486
};
\addlegendentry{L=300}

\end{axis}

\end{tikzpicture}}}
\end{minipage}
\qquad
\begin{minipage}[t]{0.45\textwidth}
\vspace{0pt} 
{\resizebox{1\textwidth}{!}{% This file was created by tikzplotlib v0.9.6.
\begin{tikzpicture}

\definecolor{color0}{rgb}{1,0.549019607843137,0}

\begin{axis}[
legend cell align={left},
legend style={fill opacity=1, draw opacity=1, text opacity=1, draw=white!80!black},
tick align=outside,
tick pos=left,
x grid style={white!69.0196078431373!black},
xmin=0.5, xmax=4,
xtick style={color=black},
xtick={1.25,2.25,3.25},
xticklabels={
{$\begin{array}{rl} P&=0.5 \\ L&=300\end{array}$},
{$\begin{array}{rl} P&=0.5 \\ L&=200\end{array}$},
{$\begin{array}{rl} P&=0.7 \\ L&=200\end{array}$}},
y grid style={white!69.0196078431373!black},
ylabel={Percentage of masked entries of $\mathcal{M}$},
ymin=0.115470886230469, ymax=22.4332046508789,
ytick style={color=black}
]
\addplot [blue!54.5098039215686!black, forget plot]
table {%
1 7.26699829101562
1 3.1036376953125
};
\addplot [blue!54.5098039215686!black, forget plot]
table {%
1 12.3970031738281
1 18.6149597167969
};
\addplot [blue!54.5098039215686!black, forget plot]
table {%
0.925 3.1036376953125
1.075 3.1036376953125
};
\addplot [blue!54.5098039215686!black, forget plot]
table {%
0.925 18.6149597167969
1.075 18.6149597167969
};
\addplot [blue!54.5098039215686!black, mark=*, mark size=3, mark options={solid,fill opacity=0}, only marks, forget plot]
table {%
1 20.64208984375
};
\addplot [blue!54.5098039215686!black, forget plot]
table {%
2 4.949951171875
2 2.16751098632812
};
\addplot [blue!54.5098039215686!black, forget plot]
table {%
2 8.51097106933594
2 11.5699768066406
};
\addplot [blue!54.5098039215686!black, forget plot]
table {%
1.925 2.16751098632812
2.075 2.16751098632812
};
\addplot [blue!54.5098039215686!black, forget plot]
table {%
1.925 11.5699768066406
2.075 11.5699768066406
};
\addplot [blue!54.5098039215686!black, mark=*, mark size=3, mark options={solid,fill opacity=0}, only marks, forget plot]
table {%
2 14.0312194824219
};
\addplot [blue!54.5098039215686!black, forget plot]
table {%
3 2.8076171875
3 1.12991333007812
};
\addplot [blue!54.5098039215686!black, forget plot]
table {%
3 4.80804443359375
3 6.74819946289062
};
\addplot [blue!54.5098039215686!black, forget plot]
table {%
2.925 1.12991333007812
3.075 1.12991333007812
};
\addplot [blue!54.5098039215686!black, forget plot]
table {%
2.925 6.74819946289062
3.075 6.74819946289062
};
\addplot [color0, forget plot]
table {%
1.5 11.3550186157227
1.5 3.52096557617188
};
\addplot [color0, forget plot]
table {%
1.5 16.6057586669922
1.5 21.4187622070312
};
\addplot [color0, forget plot]
table {%
1.425 3.52096557617188
1.575 3.52096557617188
};
\addplot [color0, forget plot]
table {%
1.425 21.4187622070312
1.575 21.4187622070312
};
\addplot [color0, mark=*, mark size=3, mark options={solid,fill opacity=0}, only marks, forget plot]
table {%
1.5 3.22418212890625
};
\addplot [color0, forget plot]
table {%
2.5 7.67726898193359
2.5 2.44064331054688
};
\addplot [color0, forget plot]
table {%
2.5 11.2514495849609
2.5 15.9957885742188
};
\addplot [color0, forget plot]
table {%
2.425 2.44064331054688
2.575 2.44064331054688
};
\addplot [color0, forget plot]
table {%
2.425 15.9957885742188
2.575 15.9957885742188
};
\addplot [color0, mark=*, mark size=3, mark options={solid,fill opacity=0}, only marks, forget plot]
table {%
2.5 2.23312377929688
};
\addplot [color0, forget plot]
table {%
3.5 4.23908233642578
3.5 1.28860473632812
};
\addplot [color0, forget plot]
table {%
3.5 6.23855590820312
3.5 9.20181274414062
};
\addplot [color0, forget plot]
table {%
3.425 1.28860473632812
3.575 1.28860473632812
};
\addplot [color0, forget plot]
table {%
3.425 9.20181274414062
3.575 9.20181274414062
};
\addplot [color0, mark=*, mark size=3, mark options={solid,fill opacity=0}, only marks, forget plot]
table {%
3.5 1.18637084960938
3.5 9.63897705078125
3.5 9.3475341796875
3.5 9.50546264648438
3.5 9.393310546875
};
\path [draw=blue!54.5098039215686!black, fill=blue!54.5098039215686!black]
(axis cs:0.85,7.26699829101562)
--(axis cs:1.15,7.26699829101562)
--(axis cs:1.15,9.40042299582862)
--(axis cs:1.075,9.69161987304688)
--(axis cs:1.15,9.98281675026513)
--(axis cs:1.15,12.3970031738281)
--(axis cs:0.85,12.3970031738281)
--(axis cs:0.85,9.98281675026513)
--(axis cs:0.925,9.69161987304688)
--(axis cs:0.85,9.40042299582862)
--(axis cs:0.85,7.26699829101562)
--cycle;
\path [draw=blue!54.5098039215686!black, fill=blue!54.5098039215686!black]
(axis cs:1.85,4.949951171875)
--(axis cs:2.15,4.949951171875)
--(axis cs:2.15,6.32631993446697)
--(axis cs:2.075,6.53839111328125)
--(axis cs:2.15,6.75046229209553)
--(axis cs:2.15,8.51097106933594)
--(axis cs:1.85,8.51097106933594)
--(axis cs:1.85,6.75046229209553)
--(axis cs:1.925,6.53839111328125)
--(axis cs:1.85,6.32631993446697)
--(axis cs:1.85,4.949951171875)
--cycle;
\path [draw=blue!54.5098039215686!black, fill=blue!54.5098039215686!black]
(axis cs:2.85,2.8076171875)
--(axis cs:3.15,2.8076171875)
--(axis cs:3.15,3.54399583162956)
--(axis cs:3.075,3.76663208007812)
--(axis cs:3.15,3.98926832852669)
--(axis cs:3.15,4.80804443359375)
--(axis cs:2.85,4.80804443359375)
--(axis cs:2.85,3.98926832852669)
--(axis cs:2.925,3.76663208007812)
--(axis cs:2.85,3.54399583162956)
--(axis cs:2.85,2.8076171875)
--cycle;
\path [draw=color0, fill=color0]
(axis cs:1.35,11.3550186157227)
--(axis cs:1.65,11.3550186157227)
--(axis cs:1.65,14.0385623419551)
--(axis cs:1.575,14.29443359375)
--(axis cs:1.65,14.5503048455449)
--(axis cs:1.65,16.6057586669922)
--(axis cs:1.35,16.6057586669922)
--(axis cs:1.35,14.5503048455449)
--(axis cs:1.425,14.29443359375)
--(axis cs:1.35,14.0385623419551)
--(axis cs:1.35,11.3550186157227)
--cycle;
\path [draw=color0, fill=color0]
(axis cs:2.35,7.67726898193359)
--(axis cs:2.65,7.67726898193359)
--(axis cs:2.65,9.5693281046049)
--(axis cs:2.575,9.74349975585938)
--(axis cs:2.65,9.91767140711385)
--(axis cs:2.65,11.2514495849609)
--(axis cs:2.35,11.2514495849609)
--(axis cs:2.35,9.91767140711385)
--(axis cs:2.425,9.74349975585938)
--(axis cs:2.35,9.5693281046049)
--(axis cs:2.35,7.67726898193359)
--cycle;
\path [draw=color0, fill=color0]
(axis cs:3.35,4.23908233642578)
--(axis cs:3.65,4.23908233642578)
--(axis cs:3.65,5.22750054192949)
--(axis cs:3.575,5.32493591308594)
--(axis cs:3.65,5.42237128424238)
--(axis cs:3.65,6.23855590820312)
--(axis cs:3.35,6.23855590820312)
--(axis cs:3.35,5.42237128424238)
--(axis cs:3.425,5.32493591308594)
--(axis cs:3.35,5.22750054192949)
--(axis cs:3.35,4.23908233642578)
--cycle;
\addplot [blue!54.5098039215686!black, forget plot]
table {%
0.925 9.69161987304688
1.075 9.69161987304688
};
\addplot [blue!54.5098039215686!black, forget plot]
table {%
1.925 6.53839111328125
2.075 6.53839111328125
};
\addplot [blue!54.5098039215686!black]
table {%
2.925 3.76663208007812
3.075 3.76663208007812
};
\addlegendentry{healthy}
\addplot [color0, forget plot]
table {%
1.425 14.29443359375
1.575 14.29443359375
};
\addplot [color0, forget plot]
table {%
2.425 9.74349975585938
2.575 9.74349975585938
};
\addplot [color0]
table {%
3.425 5.32493591308594
3.575 5.32493591308594
};
\addlegendentry{diseased}
\end{axis}

\end{tikzpicture}}}
\end{minipage}
\label{graphs}
	\caption{On the left, we present Dice scores on the BRATS2020 test set for different noise levels $L$ and probability thresholds $P$. On the right, we show the percentage of the masked entries of $\mathcal{M}$ for the healthy (blue) and diseased (orange) groups for exemplary settings of $P$ and $L$.} 
 \label{fig:hyper}
\end{figure}
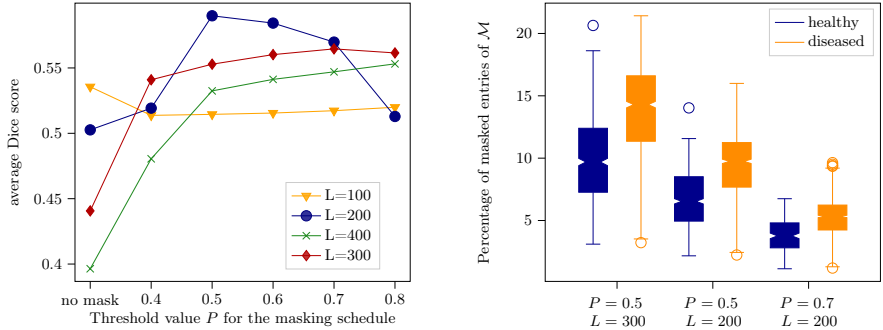

Taking advantage of the binary nature of Bernoulli diffusion models, we can directly extract image-level anomaly scores from the model output. In Fig. \ref{fig:hyper} (right), we plot the percentage of masked entries in $\mathcal{M}$ for different masking thresholds $P$ and noise levels $L$. We plot these values separately for the BRATS2020 test set of diseased and healthy slices. The box plots show clear differences between the two groups, indicating that the masked flipping probabilities extracted by our masking schedule are indicative for out-of-distribution data.

\subsubsection{Comparison to the State of the Art}
\begin{figure}[t]
    \centering
    \includegraphics[width=\textwidth]{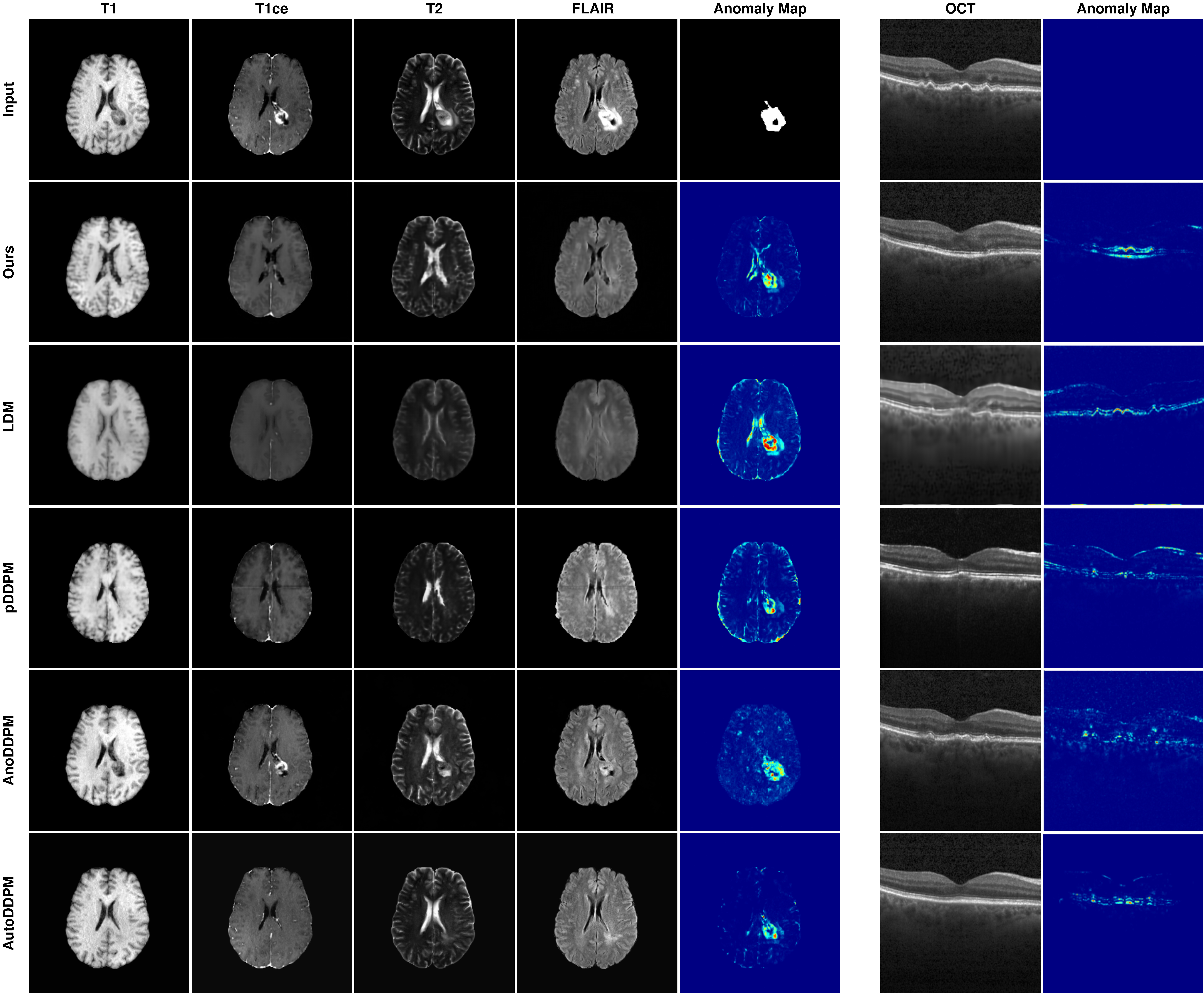}
    \caption{Qualitiative results of all comparing methods for a subject from the BRATS2020 test set, as well as a subject from the OCT2017 dataset suffering from drusen.}
    \label{fig:qualiresults}
\end{figure}

 We compare our method to \textit{AnoDDPM} \cite{wyatt2022anoddpm}, \textit{pDDPM} \cite{behrendt2024patched}, a latent diffusion model \textit{LDM} \cite{pinaya2022fast}, and \textit{AutoDDPM} \cite{bercea2023mask}. Implementation details can be found at \url{https://github.com/JuliaWolleb/Anomaly_berdiff}.
 Quantitative results on the BRATS2020 test set are summarized in Table \ref{tab:res_brats}, including the Dice scores, the area under the precision-recall curve (AUPRC), the peak signal-to-noise ratio (PSNR) between $x$ and $\hat{x}$, as well as the sampling time and GPU-memory consumption per input image. Our method achieves anomaly detection scores comparable to the state of the art, while significantly reducing the sampling time to 5\si{\second}. 
Compared to the \textit{LDM}, which also applies a masking schedule in a compressed latent space, we obtain higher scores, highlighting the effectiveness of our proposed binary masking scheme.
We present exemplary results for all methods in Fig. \ref{fig:qualiresults} for a qualitative comparison.
\begin{table}[b]
\caption{We report the mean $\pm$ standard deviation over the  BRATS2020 test set of diseased slices for the different comparing methods.}\label{tab:res_brats}
%\begin{tabular}{l|cccccc}
%\begin{tabular}{l|c{3cm}  c{3cm}  c{3cm} c{3cm} c{3cm} c{3cm}}
\begin{tabular}{l|p{1.4cm} p{1.9cm} p{1.9cm} p{1.6cm}  >{\centering}p{1cm} p{1.2cm} }
\hline
Method &  Noise & Dice  $\uparrow$ & AUPRC  $\uparrow$ & PSNR  $\uparrow$ & Time $\downarrow$ & Mem. $\downarrow$ \\
\hline
AnoDDPM         & Simplex   &$0.594 \pm 0.26$ & $0.727 \pm 0.28$ & $25.45 \pm 2.4$ & \hphantom{0}50\si{\second} &  \qty{3.85}{\giga\byte}  \\
pDDPM           & Simplex  &$0.473 \pm 0.20$ & $0.471 \pm 0.24$ & $22.82 \pm 1.9$ &  103\si{\second}  & \qty{3.44}{\giga\byte} \\
LDM             & Gaussian  &$0.490 \pm 0.24$ & $0.576 \pm 0.27$ & $20.03 \pm 2.2$ & \hphantom{00}8\si{\second} & \qty{1.87}{\giga\byte}  \\
AutoDDPM        & Gaussian   &$0.568 \pm 0.21$ & $0.632 \pm 0.27$ & $26.06 \pm 1.3$ & \hphantom{0}32\si{\second} &  \qty{4.26}{\giga\byte} \\
%\hdashline
Ours (unmasked) & Bernoulli &$0.503 \pm 0.22$ & $0.515 \pm 0.26$ & $23.31 \pm 2.0$ & \hphantom{00}4\si{\second} &  \qty{1.47}{\giga\byte} \\
Ours (masked)   & Bernoulli &$0.590 \pm 0.24$ & $0.656 \pm 0.27$ & $25.63 \pm 2.1$ & \hphantom{00}5\si{\second} &  \qty{1.47}{\giga\byte} \\
\hline
\end{tabular}
\end{table}
 On the BRATS2020 dataset, all methods produce good reconstruction results.
\textit{AutoDDPM} generates high-quality restorations, also reflected in the high PSNR values.
  Our method obtains a clear anomaly map with slight changes in the ventricles.
  %We want to emphasize that when restoring a healthy subject, deformations must be reversed, resulting in high anomaly scores in the deformed areas that are not captured by the ground truth tumor masks.
 As noted by \cite{meissen2021challenging}, hyperintense lesions can be effectively identified using a simple thresholding function. However, addressing abnormalities in anatomical structure deformations requires more than just eliminating hyperintense tissue. 
 Thus, we evaluate our method for drusen detection in OCT images. As shown in Fig. 4 on the right, our approach successfully restores a smooth RPE. Additional examples are provided in the Supplementary Material.

\section{Conclusion}
In this paper, we present a novel unsupervised anomaly detection method based on a latent Bernoulli diffusion model. By exploiting the binary nature of Bernoulli noise, we can extract anomaly scores directly from the predicted flipping probabilities. Based on this assumption, we design a novel masking procedure to improve the pixel-wise anomaly detection performance in medical images.
We evaluate our approach on two different medical datasets and successfully translate images containing anomalies into images without, resulting in competitive anomaly scores. Quantitative comparisons show that our method significantly reduces sampling time and memory consumption while still providing convincing restoration of healthy subjects.
Our approach is particularly promising for 3D applications, where memory constraints and sampling times are significant challenges. In addition, this work creates opportunities for other binary downstream tasks in medical image analysis.  For future work, an extension to binary convolutional neural networks \cite{lin2017towards} can be explored.

 \bibliographystyle{splncs04}
 \bibliography{references}

\end{document}

\title{Supplementary Material}

\author{}

\institute{}

\maketitle              % typeset the header of the contribution
\section{Bernoulli Posterior Distribution}

\begin{equation}
\label{eq:rep0}
\theta_{\text{post}}(z_t, z_0)=  \frac{[(1-\beta_t)z_t + 0.5\beta_t] \odot [\bar{\alpha_t} z_0 + 0.5b_t]}{\mathbf{Z}} ,
\end{equation}
where 
\begin{equation}
\begin{aligned}
    \mathbf{Z} &= [(1-\beta_t)z_t + 0.5\beta_t] \odot [\bar{\alpha_t} z_0 + 0.5b_t] \\
    &+ [(1-\beta_t)(1-z_t) + 0.5\beta_t] \odot [\bar{\alpha}_t (1-\epsilon_\theta(z_t, t)) + 0.5b_t],\\
&b_t= %(1 - \beta^t)
(1 - \beta_t)
b_{t-1} + 0.5 \beta_t, \quad \text{and} \quad b_1 = 0.5 \beta_1.
\end{aligned}
\label{eq:rep0}
\end{equation}

\section{Further Qualitative Anomaly Detection Results}

\begin{figure}[h]
    \centering
    \includegraphics[width=\textwidth]{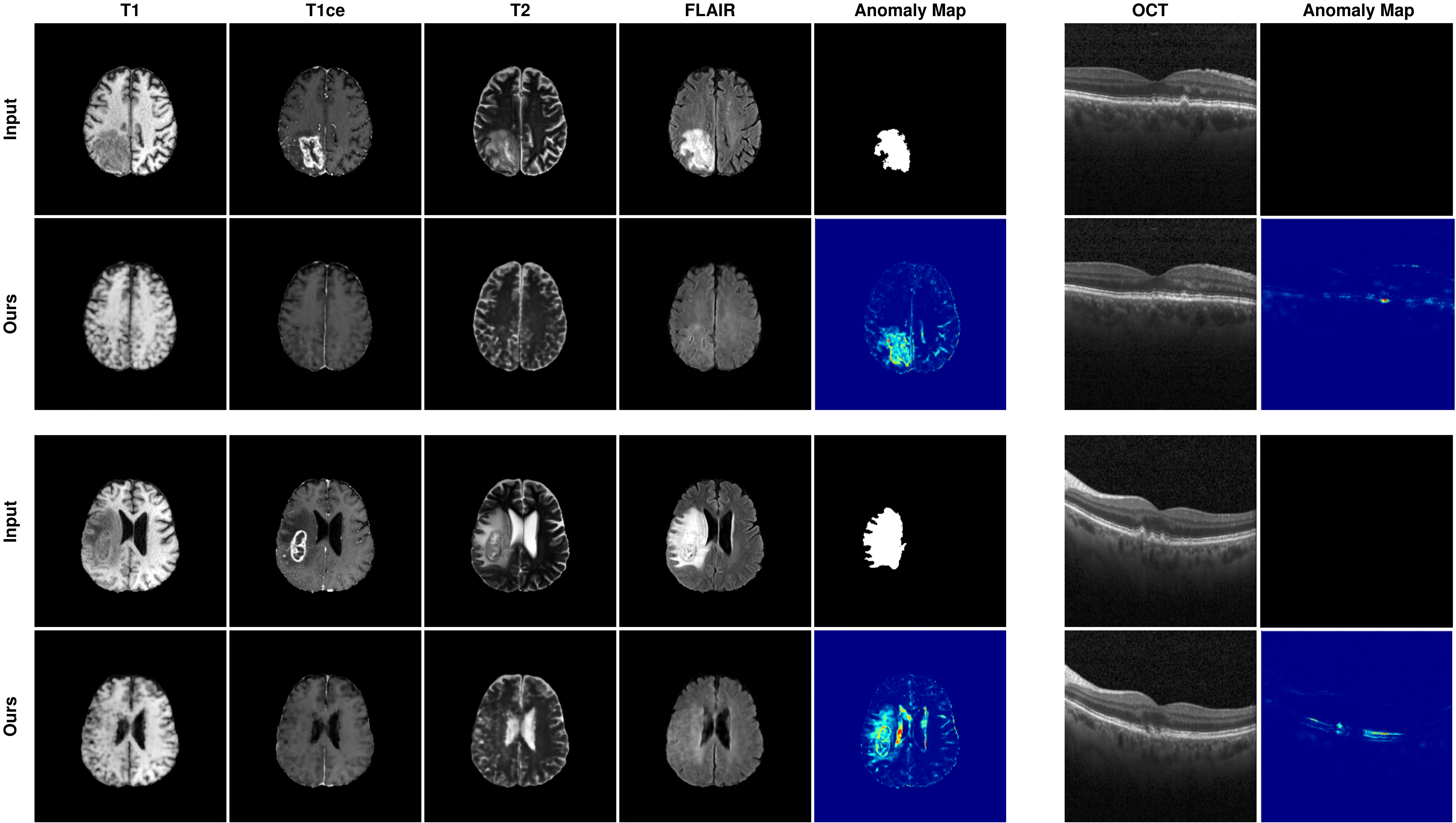}
    \caption{We provide more qualitative results of our method for two subjects of the BRATS2020 test set, as well as two subjects of the OCT2017 test set. Here, we choose $P=0.5$ and $L=300$.}
    \label{fig:qualiresults}
\end{figure} 

\section{Qualitative Hyperparameter Analysis}
\begin{figure}[H]
\includegraphics[width=0.85\textwidth]{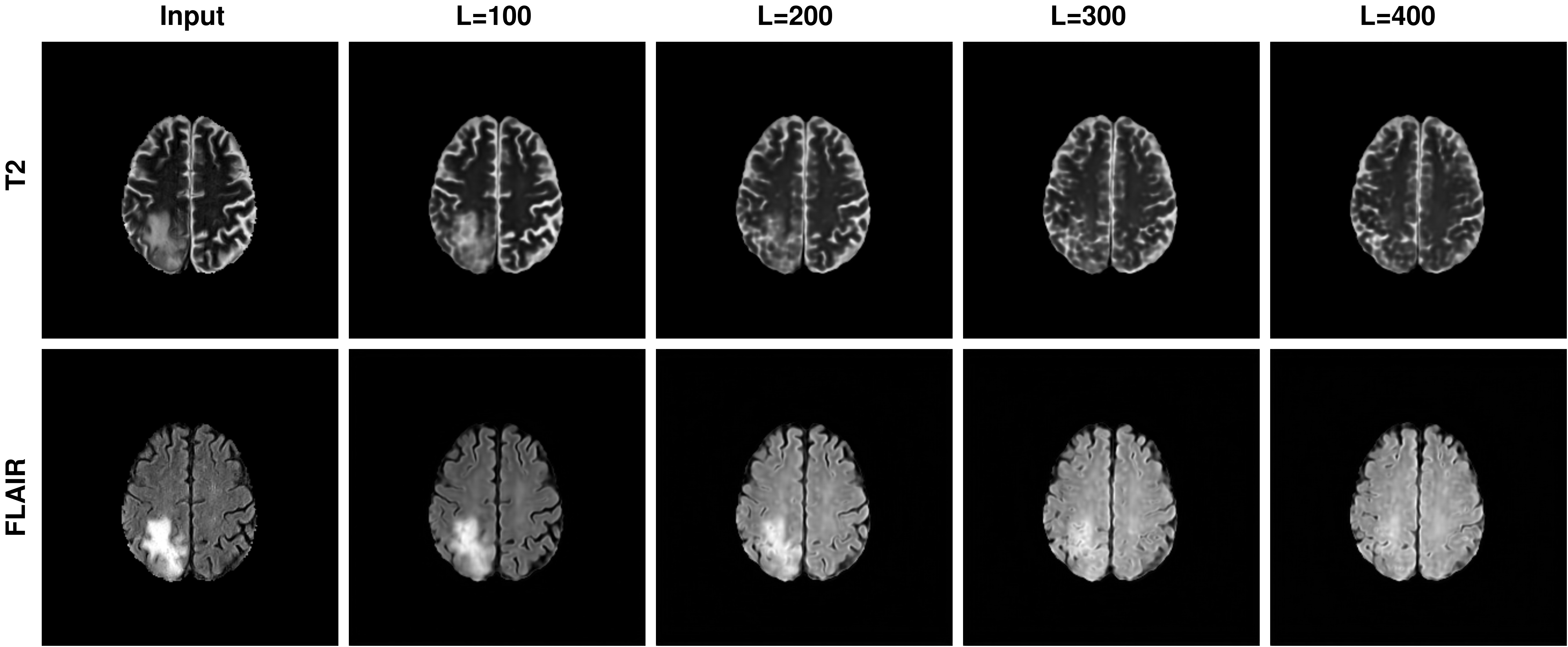}
\caption{With a fixed probability threshold of $P=0.5$, we demonstrate the impact of increasing noise levels $L$ on T2-weighted as well as FLAIR MR images.}\label{timeline} 
\end{figure}

\includegraphics[width=\textwidth]{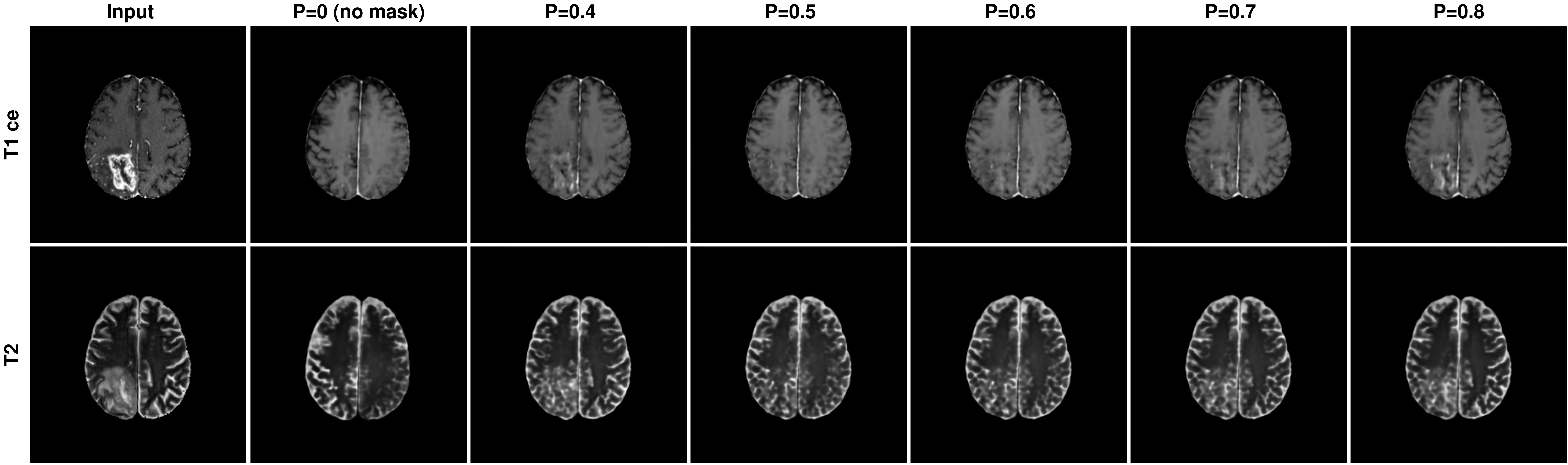}
\captionof{figure}{
Inference is performed for increasing probability thresholds 
$P$ for T2-weighted and 
contrast enhanced T1-weighted MR images of the BRATS2020 test set, with a fixed $L=300$.
} \label{timeline}

\section{Inference on a Healthy Sample}

\includegraphics[width=0.8\textwidth]{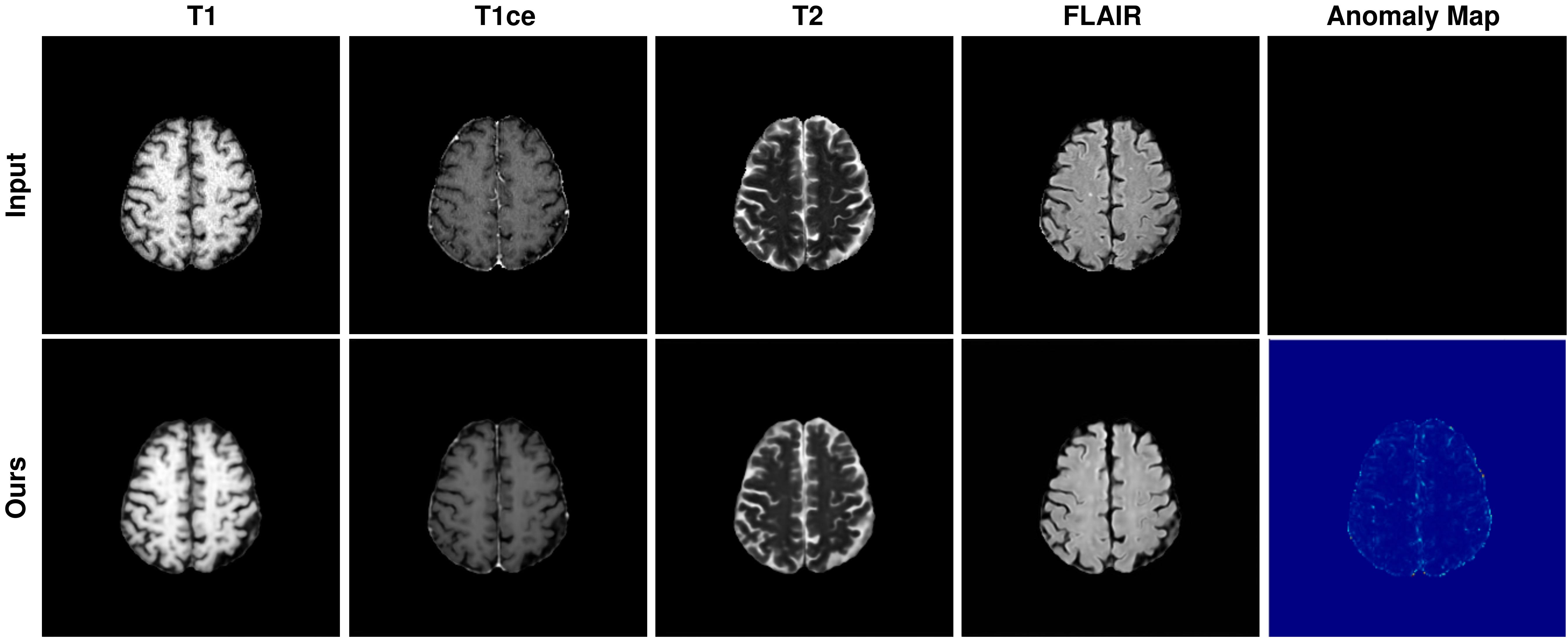}
\captionof{figure}{We run inference on a healthy slice of the BRATS2020 test set. The reconstruction is close to the input, resulting in an anomaly map close to zero.} \label{timeline}